\PassOptionsToPackage{table}{xcolor}
\documentclass[sigconf]{acmart}
\AtBeginDocument{%
  }

\setcopyright{acmlicensed}
\copyrightyear{2027}
\acmYear{2027}
\acmDOI{}
\acmISBN{}

\usepackage{multirow}
\usepackage{colortbl}
\usepackage{pgfplots}
\usepackage{algorithm}
\usepackage{algorithmic}
\usepackage{listings}
\pgfplotsset{compat=1.18}

\begin{document}

\title{Constraint Dependency Graph Retrieval for Compliance Detection on Long SKILL Documents}

\author{Shuaitao Zhao}
\authornotemark[1]
\email{zhaoshuaitao@meituan.com}
\affiliation{\institution{Meituan}\city{Beijing}\country{China}}

\author{Feng Ni}

\authornote{Equal contribution.}
\email{nifeng02@meituan.com}
\affiliation{\institution{Meituan}\city{Beijing}\country{China}}

\author{Lichao Ma}
\affiliation{\institution{Peking University}\city{Beijing}\country{China}}

\author{Yuchen Wei}
\affiliation{\institution{Zhejiang University}\city{Hangzhou}\country{China}}

\author{Shuaike Li}
\affiliation{\institution{University of Science and Technology of China}\city{Hefei}\country{China}}

\author{Jiaye Lin}
\affiliation{\institution{Meituan}\city{Beijing}\country{China}}

\author{Fei Han}
\affiliation{\institution{Meituan}\city{Beijing}\country{China}}

\author{Yang Wei}
\affiliation{\institution{Meituan}\city{Beijing}\country{China}}

\author{Lu Pan}
\affiliation{\institution{Meituan}\city{Beijing}\country{China}}

\renewcommand{\shortauthors}{}

\begin{abstract}
Large Language Model (LLM) agents carry out business operations under enterprise documents we refer to as SKILL documents, which take the form of standard operating procedures (SOPs) that state the conditions to verify and the actions to take before an operation proceeds.
Long, densely cross-referenced SKILL documents exceed the context budget of small deployed auditors. Similarity-based retrieval then brings in irrelevant rules while omitting mandatory prerequisites, and the two-stage audit policy provides no trainable signal for the retrieval stage.
We present SkillCDG, a compliance detection framework that reformulates each SKILL document as a constraint dependency graph, shifting rule retrieval from similarity-based matching to closure-based selection under the deployment's context budget.
Each atomic rule is anchored to its source text span as a graph node, and mandatory prerequisite relations form the sole edge type. Policy-generated retrieval keywords activate seed rules, whose dependency closure yields a compact and conditionally complete rule context within that budget.
The same graph also supports a joint optimization scheme. The retrieval stage is trained with group-relative policy optimization using a verifiable reward that combines output-format validity with downstream verdict correctness, with closure non-emptiness serving as a validity gate, while selective on-policy distillation concentrates teacher supervision on high-entropy reasoning tokens, with verdict and termination tokens always included, to improve verdict accuracy.
Experiments on three enterprise datasets and two public benchmarks show that SkillCDG improves accuracy and F1 over the full-document and retrieval-based baselines.
\end{abstract}




\maketitle

\section{Introduction}
\label{sec:intro}

\begin{figure}[h]
\centering
\includegraphics[width=\linewidth]{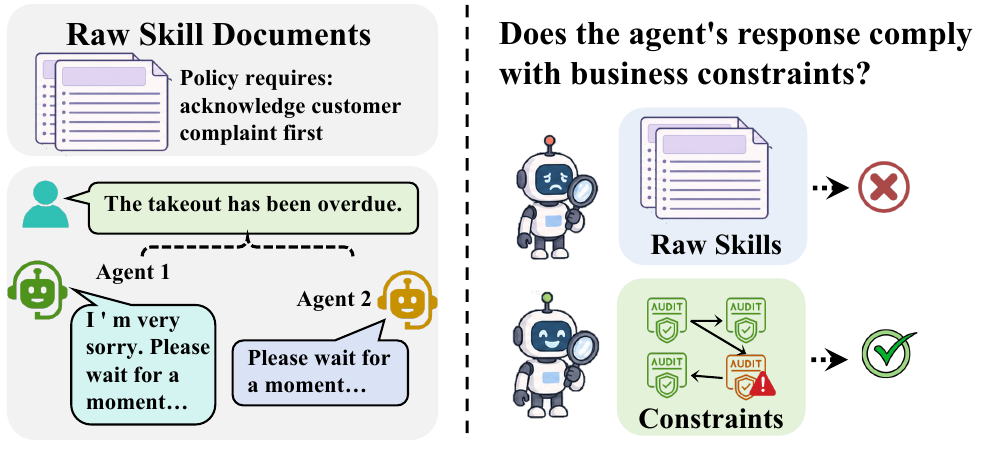}
\caption{Identical responses may result from compliant and non-compliant procedures. Reliable auditing requires the applicable rule and its complete set of policy prerequisites.}
\Description{Motivation of SkillCDG}
\label{fig:intro}
\end{figure}

LLM agents increasingly carry out business actions under
written enterprise policies, which we refer to as SKILL documents and which state
the conditions to verify and the actions to take before an operation proceeds
\citep{yao2024tau,barres2025tau,diao2025guidebench,qi2025agentif}. Because such
a document is a specification rather than a preference, the same visible action
can arise from a procedure that was followed and from one that was not, and in
domains such as billing, refunds and access control this difference persists even
when the outcome itself looks correct, as Figure~\ref{fig:intro} illustrates. This makes the document an object of audit
as well as a guide, and it defines the task we study here, compliance detection,
which is to take an interaction together with the response that followed it and
produce a verdict, identifying the violated rule and its source span whenever the
verdict is non-compliant. Recent work treats this task as a distinct target, and
CompliBench provides an explicit benchmark for it
\citep{wen2026policyguard,mishra2026probe,yang2026complibench}.

The difficulty begins with scale, since enterprise policies grow as the business
grows and a single document can exceed ten thousand tokens
\citep{liu2026policy}. Rules inside such a document also refer to one another, so
the rules that govern one scenario are rarely self-contained and are assembled
from several places in the text. The natural response is to place the whole
document in the auditor's context, yet a small deployed auditor has room for
only a fraction of it, and that fraction is spent largely on rules the current
scenario does not activate. A more economical response is to retrieve the rules
that appear relevant, and much of the recent progress in retrieval has taken this
form \citep{lewis2020rag}, which shifts the burden from reading to locating, since
the auditor is now responsible for finding the governing rules before judging
anything. That direction has produced real gains, and it was our starting
point, which raises the question of whether relevance is the target worth
pursuing.

A verdict depends on the set of rules that are jointly required, and not on the
set of rules that read as similar to the scenario. When one rule requires another
to be judged first, retrieving the first without the second leaves the verdict
unsupported even though the retrieved rule is plainly relevant, and in the other
direction a prerequisite may be phrased in words that the scenario does not use,
so no degree of similarity will surface it. Similarity therefore answers a
question that is close to, but not the same as, the one the verdict poses, and
this is the first of the two gaps we address. Recent work has moved in a related
direction by selecting passage sets that satisfy a query as a whole
\citep{lee2025setr} and by showing that coverage-based retrieval metrics correlate more strongly with
downstream RAG information coverage than relevance-based metrics
\citep{samuel2026beyondrelevance}, while graph-based retrieval has improved
multi-hop access by following semantic relations
\citep{edge2024graphrag,guo2024lightrag,gutierrez2024hipporag}. These
approaches either optimize a retrieval score or measure evidence use after retrieval. Neither certifies that no prerequisite was omitted, and
a longer context window is no substitute, since dispersed evidence is often left
unused even when it is present \citep{liu2024lost,du2025context} and compression
risks discarding the one prerequisite that changes the verdict
\citep{jiang2023longllmlingua,pan2024llmlingua2}.

Auditing is not a single decision either, since it proceeds in two steps, the
first selecting the rules to be read and the second reaching a verdict. The retrieval stage produces keywords rather than an answer and leaves no record of which rules
should have been selected, so a correct verdict does not show that the right rules
were retrieved and an incorrect one does not show which of the two stages went astray.
Group-relative policy optimization offers a way to train the retrieval stage without a
separate value model \citep{shao2024deepseekmath}, and related work has used
outcome reward to shape an intermediate search decision
\citep{jin2025searchr1,zhang2026tosearch}, yet two problems remain.
When all samples in a group receive the same reward, the relative
advantage is zero and no update follows
\citep{yu2025dapo,he2026advantagecollapse}, and where finer credit is wanted,
existing agents rely on repeated environment states or on annotated evidence
\citep{feng2025gigpo}, and the retrieval stage provides neither form of
retrieval-level supervision. A teacher can
supply the missing signal, yet on-policy distillation applied across every token
is costly \citep{gu2024minillm,agarwal2024gkd} and a small share of tokens appears
to decide an outcome \citep{wang2025beyond8020}.

This paper presents SkillCDG, which compiles each SKILL document into a constraint
dependency graph and audits against the graph rather than against the raw
document. Each node is a single atomic rule recorded together with the span of the
source document it came from, and the graph carries one kind of edge, written
\texttt{require}, which holds between two rules when the second is required to be
judged before the first. At inference the scenario activates a small set of seed
rules, and the rules that are read are those seeds together with everything
reachable from them along \texttt{require} edges, so the resulting set is small,
conditionally complete with respect to the activated seeds and the extracted
\texttt{require} edges, and traceable rule by rule to the original text. The same
graph also supplies the non-emptiness half of the validity check: a selection is
valid when it parses and reaches at least one rule, and non-emptiness needs no
retrieval-level labels, learned reward model, or separate value model. On that basis the two steps are trained together. The retrieval stage
is trained by group-relative policy optimization with a verifiable reward that
combines output-format validity with downstream verdict correctness, and the
judgment stage is sharpened by an on-policy distillation in which the teacher
scores the student's own tokens rather than generating its own, with supervision
concentrated on the highest-entropy reasoning tokens, while verdict and
termination tokens are always included.

We evaluate SkillCDG on three enterprise datasets and two public benchmarks.
Accuracy (ACC) improves over the full-document and retrieval-based baselines,
and F1 also rises.
Most of the gain over frozen-retrieval baselines comes from training the retrieval
stage, while the closure itself is the largest single contributor. An ablation that
removes the dependency closure, freezes retrieval, removes the entropy gate, and
removes the teacher separates the contribution of each component.

This work makes three contributions. (1) We compile each SKILL document
into a constraint dependency graph whose nodes are atomic rules anchored to spans
of the source text and whose only edge, \texttt{require}, marks a mandatory
prerequisite, and we use the dependency closure of scenario-activated seed rules
as the judgment context, which is tight, conditionally complete with respect to
the activated seeds and the extracted \texttt{require} edges, and traceable to the
original text. (2) We treat the retrieval stage and the judgment stage as two
stages of one policy and train them jointly, training the retrieval stage by
group-relative policy optimization with a verifiable reward that combines
output-format validity with downstream verdict correctness, with closure
non-emptiness serving as a validity gate, and relying on no retrieval-level labels,
no reward model and no critic. Within the same optimization the judgment stage is
sharpened by an entropy-gated selective on-policy distillation in which the teacher
scores rather than generates, so that supervision is concentrated on the
highest-entropy reasoning tokens, with verdict and termination tokens always
included. (3) We validate the method on three enterprise datasets and two public
benchmarks, and it improves accuracy and F1.

\section{Related Work}
\label{sec:related}

\subsection{Retrieval-Augmented Generation}

Retrieval-augmented generation supplements a language model's parametric knowledge with evidence retrieved at inference time~\cite{lewis2020rag}.
The original framework retrieves passages by dense similarity~\cite{karpukhin2020dpr} and prepends them to the prompt, making similarity the default retrieval objective for this line of work.
Subsequent research refines retrieval along this axis: \citet{lee2025setr} shifts from ranking to set selection to better cover multi-facet queries, while \citet{samuel2026beyondrelevance} shows that coverage-based retrieval metrics correlate more strongly with downstream RAG information coverage than relevance-based metrics.
Work on the consumption side finds that long contexts do not guarantee faithful use of retrieved evidence~\cite{liu2024lost,du2025context,bai2024longbench}, which motivates prompt-compression methods~\cite{jiang2023longllmlingua,pan2024llmlingua2} that trim by information density.
Graph-based methods restructure the corpus before retrieval: GraphRAG~\cite{edge2024graphrag} builds a knowledge graph from the corpus and aggregates community-level summaries for query-focused tasks, while LightRAG~\cite{guo2024lightrag} and HippoRAG~\cite{gutierrez2024hipporag} add lightweight graph indices or biologically inspired associations for cheaper graph-level access.
These methods move retrieval from passages to graph structures, but the edges remain semantic associations and the retrieval objective stays relevance-maximizing.
They do not model mandatory prerequisites between rules or compute a closure over them, so they do not provide the completeness guarantee that compliance reasoning requires.
Our work adopts the graph representation but replaces the edge semantics with a single mandatory-dependency type and the retrieval objective with dependency closure solving.

\subsection{Instruction Following and Compliance Detection}
Instruction-following evaluation in chat settings is established by \citet{zhou2023instruction}. Earlier work studied learning from human feedback as a way to align model outputs with stated preferences~\cite{ziegler2019rlhf}.
In enterprise settings, evaluation has shifted toward agentic and policy-bound tasks: AgentIF~\cite{qi2025agentif} benchmarks instruction following across real agentic applications with long and multi-constraint instructions, GuideBench~\cite{diao2025guidebench} evaluates domain-oriented guideline adherence, and $\tau$-bench~\cite{yao2024tau} and its dual-control extension~\cite{barres2025tau} supply interactive tool-agent-user environments.
Tool-augmented agents~\cite{yao2023react,schick2024toolformer} and chain-of-thought prompting~\cite{wei2022chain} broaden the range of tasks and reasoning traces that such evaluations must cover.
These benchmarks measure end-to-end compliance but keep retrieval, dependency resolution, and verdict production inside the model, so a failure is not attributed to a specific stage.
Recent work begins to isolate the compliance check from the agent under audit: \citet{liu2026policy} studies how agents internalize complex policy documents, PolicyGuard~\cite{wen2026policyguard} learns efficient guardrails for compliance, PolicyGuide~\cite{kang2026policyguide} extends policy compliance to full workflow guidance, PROBE~\cite{mishra2026probe} trains an auditor agent to verify tool-using agents against policy, and CompliBench~\cite{yang2026complibench} benchmarks LLM-as-judge for violation detection and localization in multi-turn dialogues.
\citet{rachmil2026whitening} take a training-free approach and detect policy violations by comparing hidden activation norms against in-policy statistics.
Most of these works take the policy document as given and leave rule selection to the judge. PolicyGuide is the exception: it compiles the policy into a workflow graph and verifies at user-turn boundaries.
Our work makes rule selection explicit: the constraint dependency graph turns the applicable-rule set into a checkable artifact, and the closure property makes completeness a stated property of the input rather than an assumption about the model.

\subsection{Reinforcement Learning and On-Policy Distillation}
Group Relative Policy Optimization (GRPO)~\cite{shao2024deepseekmath} trains language models with outcome-level rewards by normalizing advantages within a sampled group, removing the value network required by PPO-style methods~\cite{schulman2017ppo}.
Follow-up work studies its failure modes and extensions: advantage collapse~\cite{he2026advantagecollapse} diagnoses and mitigates gradient degradation when group rewards are uniform, Group-in-Group policy optimization~\cite{feng2025gigpo} extends the grouping structure to multi-step agent training, and DAPO~\cite{yu2025dapo} scales the recipe to larger settings.
Group-based reinforcement learning has also been applied to retrieval-augmented agents: Search-R1~\cite{jin2025searchr1} trains models to decide when to invoke a search engine, and \citet{zhang2026tosearch} aligns the search decision boundary via causal intervention.
Both verify the retrieval action against an external search engine; we verify it against the structural closure of a constraint graph, which requires no external search engine or search log.
On-policy distillation~\cite{gu2024minillm,agarwal2024gkd} lets a student learn from its own rollouts while a teacher provides token-level guidance, avoiding the exposure bias of supervised fine-tuning.
\citet{ramos2026recipe} combines GRPO with on-policy teacher anchoring for long-context reasoning and shows the two objectives are complementary.
\citet{wang2025beyond8020} further shows that a small set of high-entropy tokens drives most of the learning signal in reinforcement learning.
Our training uses the same GRPO and on-policy distillation recipe, with two differences: the retrieval action is verified by graph structure rather than an outcome reward, supplying a dense annotation-free signal for an otherwise unobservable intermediate step, and the distillation supervision is gated by old-policy entropy so that the teacher budget is concentrated on the decision points that most influence the final verdict.

%%%%%%%%% Preliminary %%%%%%%%%
\section{Preliminary}
\label{sec:preliminary}

\subsection{SKILL Documents and Compliance Detection}

A SKILL document is a written enterprise policy that prescribes the conditions an agent must verify and the actions it must take before completing a business operation~\citep{yao2024tau,diao2025guidebench}.
Let $\mathcal{J} = \{j_1, j_2, \dots, j_m\}$ denote a finite collection of SKILL documents loaded for a deployment.

A compliance detection instance is a triple $(x, \mathcal{J}_x, \hat{y})$, where $x$ is an interaction context, $\mathcal{J}_x \subseteq \mathcal{J}$ is the subset of SKILL documents relevant to that context, and $\hat{y}$ is the agent response under audit.
The task is to produce a verdict $v \in \{\text{compliant}, \text{non-compliant}\}$ together with a citation of the violated rule when $v = \text{non-compliant}$.

\subsection{Two-Stage Audit Policy}

We treat auditing as a two-stage policy $\pi_\theta$ executed on each instance $(x, \mathcal{J}_x, \hat{y})$.
In the first stage the policy produces retrieval keywords $q = \pi_\theta^{\mathrm{ret}}(x, \mathcal{J}_x, \hat{y})$, which are matched against the constraint dependency graph to obtain a rule set $\mathcal{R}_q$.
In the second stage the policy produces a verdict $v = \pi_\theta^{\mathrm{jdg}}(x, \mathcal{J}_x, \hat{y}, \mathcal{R}_q)$ conditioned on the retrieved rules.
The retrieval stage has no ground-truth annotation; only the final verdict label is observed.

\section{Method}
\label{sec:method}

\begin{figure*}[t]
\centering
\includegraphics[width=0.95\textwidth]{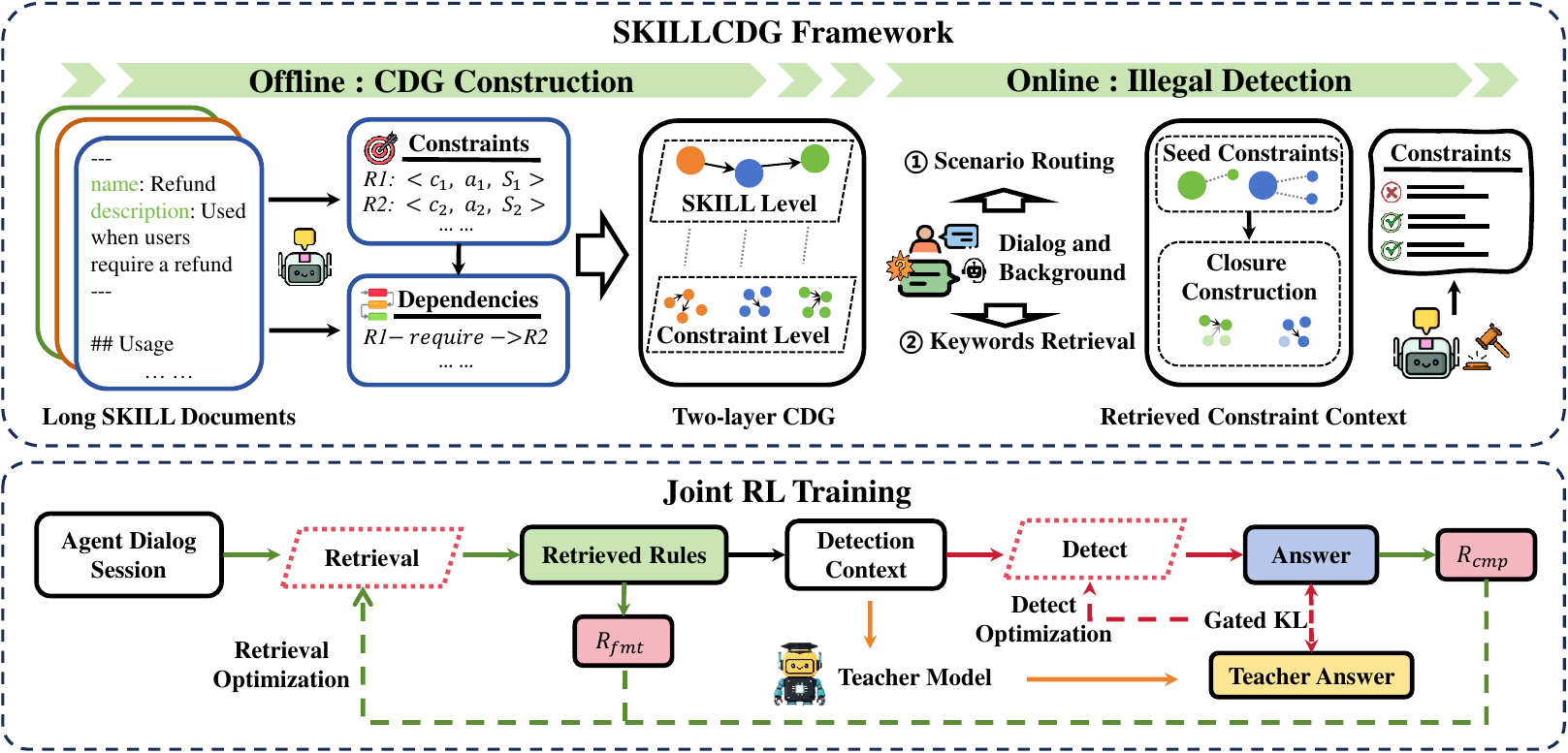}
\caption{The overview of \textsc{SkillCDG} framework.}
\Description{Overview of the SkillCDG framework, showing the constraint dependency graph construction, closure-based compliance judgment, and the joint RL training procedure.}
\label{fig:overview}
\end{figure*}

This section presents \textsc{SkillCDG}, a framework for compliance detection under long SKILL documents.
Policy documents are long and internally interdependent. Judging the full text exposes the model to rules that do not apply, while trimming by relevance can omit the prerequisites that make a rule applicable.
\textsc{SkillCDG} addresses this with two components, as illustrated in Figure~\ref{fig:overview}.

\subsection{SkillCDG Framework}
\label{sec:method-cdg}

The SkillCDG framework couples an offline compilation step with an online judgment pipeline that operates on the compiled structure.
At inference, the framework locates rules relevant to a given interaction, expands them along dependency edges into a closure, and presents the closure as a structured context from which the judgment model decides compliance.
Retrieving rules and judging them are two stages of one pipeline: the closure determines what the model sees, and the model verdict determines whether the closure was sufficient.

\subsubsection{Constraint Dependency Graph Construction}
\label{sec:method-cdg-construction}

A rule near the end of a policy document often inherits its applicability condition from a rule stated earlier, and judging the later rule in isolation leads to incorrect verdicts.
Feeding the full document to a small judgment model also fills the context with rules that have no bearing on the current scenario, which makes the applicable constraints harder to identify.
We compile each document into a constraint dependency graph that stores rules as atomic units and makes the prerequisite relations between them explicit and traversable.

Formally, for each SKILL document $j \in \mathcal{J}$, we compile a constraint dependency graph $\mathcal{G}_j = (\mathcal{R}_j, \mathcal{E}_j)$ offline, where $\mathcal{R}_j$ is a set of atomic rules and $\mathcal{E}_j$ is a set of directed edges of a single type.
An atomic rule is a triple $r = \langle c, a, \mathcal{S} \rangle$, where $c$ is an applicability condition, $a$ is a normative action, and $\mathcal{S}$ is a verbatim span of the source document that anchors the rule to its original text.
A directed edge $(r_i, r_k) \in \mathcal{E}_j$ of type \texttt{require} holds when judging the compliance of $r_i$ requires that $r_k$ be judged first.
Using a single edge type keeps the dependency semantics unambiguous: every edge is a mandatory prerequisite, and the closure operation on these edges has a completeness guarantee.

Construction starts by splitting each document at Markdown headings into blocks; an extractor model then reads each block and identifies atomic rules and the \texttt{require} edges between them.
A fusion pass merges duplicate rules across blocks, normalizes source spans, and adds cross-block edges that the per-block pass may have missed.
The result is stored in two layers: an upper layer indexes each SKILL document by its name and description for document-level routing, and a lower layer stores the rule graph of each document.
The audit policy can therefore select the relevant documents first and operate only on their rule graphs, which keeps online computation proportional to the active subset rather than the full corpus.

The source span $\mathcal{S}$ must be verbatim for traceability, so the extractor model preserves the text exactly.
We use a model of at least 8B parameters for extraction, since smaller models tend to paraphrase the source text and cause span-lookup failures at runtime.
When a span is not found at runtime, the system records a \texttt{source\_span\_error} and drops the rule.

\subsubsection{Closure-based Compliance Judgment}
\label{sec:method-closure}

At inference, the framework receives an instance $(x, \mathcal{J}_x, \hat{y})$ comprising the interaction context, the set of active documents, and a candidate response to audit.
The policy first produces a set of keywords $q$ from the interaction context and matches them against the rules of the active documents to obtain a seed set $\mathcal{R}_q^{\mathrm{seed}} \subseteq \bigcup_{j \in \mathcal{J}_x} \mathcal{R}_j$.
The seed set is only a starting point. A seed rule may depend on prerequisites that the keywords did not surface, and judging without them leaves the verdict unsupported.

We expand the seed set into a dependency closure by iteratively adding every rule reachable along \texttt{require} edges:
\begin{equation}
\label{eq:closure}
\mathrm{Cl}(\mathcal{R}_q^{\mathrm{seed}}) = \mathcal{R}_q^{\mathrm{seed}} \cup \{ r_k \mid \exists r_i \in \mathcal{R}_q^{\mathrm{seed}},\ (r_i, r_k) \in \mathcal{E} \}^{\!*},
\end{equation}
where the Kleene closure iterates until no new rule appears.
The closure $\mathrm{Cl}(\mathcal{R}_q^{\mathrm{seed}})$ is the judgment context presented to the judgment model, and the algorithm for computing it is detailed in Appendix~\ref{app:closure}.
The construction gives two properties.
Tightness holds because the closure contains only rules activated by the seed or required by an activated rule, so rules unrelated to the scenario are excluded.
Conditional completeness holds because every mandatory prerequisite of every seed rule is included by construction, so a verdict drawn from the closure is not missing a rule the document itself marks as required.

The closure depends on the graph rather than on the retriever's scoring.
A similarity-based retriever can return the same rules, but it does not establish that no prerequisite was omitted.
The closure follows the transitivity of \texttt{require} edges, and the seed set is the only variable that determines which rules enter it.
This separation lets us train the retrieval stage in Section~\ref{sec:method-joint}: better seed selection directly improves the closure, and the completeness guarantee is preserved across such improvements.

Each rule in the closure is presented to the judgment model as a three-line block listing its applicability condition, its normative action, and its verbatim source span, which lets the model read each rule as a self-contained unit and cite the source span directly when a violation is found.
The judgment model receives the closure context together with the interaction context and the candidate response, and outputs a verdict for the instance. When the verdict is non-compliant, the reason identifies the violated rule and its source span.
If the seed set is empty or the closure yields no rules, the system reports a retrieval error and does not issue a verdict.

The closure reduces the context the judgment model must process and removes rules that do not apply.
Atomizing constraints into self-contained units lowers the load on a small model that would otherwise filter relevant rules from a long document, and excluding unrelated rules concentrates attention on the constraints that apply.

\subsection{Joint RL Training}
\label{sec:method-joint}

We train the retrieval and judgment stages of the audit policy $\pi_\theta$ jointly.
Both stages use the downstream result as a signal for the upstream decision: the retrieval stage is scored by the judgment it enables, and the judgment stage receives token-level teacher feedback.
In the retrieval stage, group-relative policy optimization (GRPO) generates multiple keyword candidates per instance and scores each by the downstream judgment, which turns retrieval quality into a reward the policy can optimize.
In the judgment stage, on-policy distillation (OPD) provides token-level guidance from a teacher model, with supervision gated to the high-entropy tokens where the student is least certain.
The two stages are optimized on the same rollout batch, so that neither stage dominates.

\subsubsection{Retrieval Optimization}
\label{sec:method-retrieval}

Retrieval quality is difficult to measure directly, since the retrieval stage produces keywords rather than an answer and the dataset provides only the final verdict label.
The downstream compliance judgment, however, reflects whether the retrieved rules were sufficient for the model to reach an accurate verdict, which makes it a natural proxy for retrieval quality.
GRPO's multiple-rollout structure provides this signal: by generating $K$ keyword candidates from the current policy, computing the dependency closure for each, running the judgment model on each closure, and aggregating the judgment performance, we obtain a reward that measures how well each retrieval served the judgment.

Each candidate $q^{(i)}$ triggers the closure procedure of Section~\ref{sec:method-closure} and yields a rule set $\mathrm{Cl}(\mathcal{R}_{q^{(i)}}^{\mathrm{seed}})$.
The judgment model then generates $M$ verdict candidates $\{v^{(i,j)}\}_{j=1}^M$ conditioned on the retrieved rules.
We define the retrieval reward as a combination of a format reward and a compliance reward:
\begin{equation}
\label{eq:ret_reward}
R_{\mathrm{ret}}(q^{(i)}) = R_{\mathrm{fmt}}(q^{(i)}) + \frac{1}{M}\sum_{j=1}^{M} R_{\mathrm{cmp}}(v^{(i,j)}),
\end{equation}
where $R_{\mathrm{fmt}}$ checks that the keyword output conforms to the expected format so that the closure can be computed, and $R_{\mathrm{cmp}}$ measures whether the judgment based on the closure of $q^{(i)}$ matches the ground-truth label.
Averaging the compliance reward over $M$ judgment rollouts reduces the variance introduced by the judgment model's own stochasticity, so that the reward reflects retrieval quality rather than one verdict sample.
The group-relative advantage centers the reward within the group:
\begin{equation}
\label{eq:grpo_adv}
\hat{A}(q^{(i)}) = R_{\mathrm{ret}}(q^{(i)}) - \frac{1}{K}\sum_{j=1}^{K} R_{\mathrm{ret}}(q^{(j)}),
\end{equation}
and the GRPO objective for the retrieval stage is
\begin{equation}
\label{eq:grpo_obj}
\mathcal{L}_{\mathrm{ret}}(\theta) = -\mathbb{E}\!\left[\sum_{i=1}^{K} \hat{A}(q^{(i)}) \log \pi_\theta(q^{(i)} \mid x, \mathcal{J}_x, \hat{y})\right].
\end{equation}
Centering the reward within the group removes the need for a value network and lets the policy learn from relative performance rather than absolute scores.
When all candidates in a group receive the same reward, the advantage is zero and no gradient is produced, since a group with no variation carries no information about which action is better.
A uniformly rewarded group therefore contributes no update. The reward must vary across candidates, and the verdict-based compliance reward provides that variation.
Within a group, candidates that retrieve relevant rules and produce accurate judgments receive positive advantage, while those that miss mandatory prerequisites or retrieve irrelevant rules receive negative advantage, so the policy is pushed toward keywords that serve the judgment well.

\subsubsection{Compliance Judgment Optimization}
\label{sec:method-judgment}

Once retrieval is trained, the remaining bottleneck is the judgment stage.
Compliance detection is a 0/1 classification at the trajectory level, which means the result reward provides only a sparse signal: most of the tokens in a judgment trajectory contribute to a reasoning chain whose correctness is not evaluated until the final verdict token.
On-policy distillation (OPD)~\citep{gu2024minillm,agarwal2024gkd} addresses this by giving the student token-level guidance from a teacher model while the student runs its own rollouts, which avoids the exposure bias of supervised fine-tuning.
The teacher scores the student's tokens rather than generating its own, which avoids the cost of teacher generation while still providing dense supervision.

Uniformly distilling every token is wasteful in this setting.
Within a judgment trajectory, most tokens carry forward a reasoning chain that the student already handles with confidence, and only a small subset at the decision points where the verdict is about to flip carry the uncertainty that a teacher signal should target.
Prior work shows that a small share of high-entropy tokens drives most of the learning signal~\citep{wang2025beyond8020}.
We gate the distillation supervision by old-policy entropy: for each trajectory, we compute the entropy of the student's previous policy on every reasoning token and select the top 20\% by entropy for distillation; verdict tokens and termination tokens are always included because they directly determine the output.
Let $\mathcal{M}(v^{(i,j)})$ denote the selected token mask for trajectory $(i,j)$.
The selective OPD objective is
\begin{equation}
\label{eq:opd_loss}
\mathcal{L}_{\mathrm{jdg}}(\theta) = \mathbb{E}\!\left[\sum_{i,j} \sum_{t \in \mathcal{M}(v^{(i,j)})} \mathrm{KL}\!\left(\pi_\theta(\cdot \mid h_t) \,\|\, \pi_{\mathrm{T}}(\cdot \mid h_t)\right)\right],
\end{equation}
where $h_t$ is the token-level context and $\pi_{\mathrm{T}}$ is the teacher scoring distribution.
The per-token log-ratio is clipped at $\pm 15$ to bound the influence of outliers.

Gating by old-policy entropy concentrates the teacher budget on the tokens where the student was least certain under the previous policy, which are the decision points that determine the verdict.
Selecting 20\% of tokens reduces teacher scoring and avoids supervising tokens the student already handles well. The student still generates full trajectories, so the forward-pass cost is unchanged.

The two losses are optimized jointly on the same rollout batch:
\begin{equation}
\label{eq:combined}
\mathcal{L}(\theta) = \mathcal{L}_{\mathrm{ret}}(\theta) + \lambda\, \mathcal{L}_{\mathrm{jdg}}(\theta),
\end{equation}
where $\lambda$ balances the two stages.
Both losses are clipped independently so that one stage does not dominate the other. The loss is normalized by region (query, reason, verdict) and by slot, with failure slots kept in the denominator so that failed trajectories still contribute.
The GRPO objective gives the retrieval stage a signal that result-only supervision does not provide, and the entropy-gated distillation gives the judgment stage dense token-level guidance.
Together they address the credit-assignment problem of training a two-stage policy on compliance labels alone.

\section{Experiments}
\label{sec:experiments}

We conduct experiments to answer four research questions:
\textbf{RQ1}: How does SkillCDG compare with baselines across datasets?
\textbf{RQ2}: How does each component contribute to the overall gain?
\textbf{RQ3}: How does SkillCDG adapt as the SKILL context length increases?
\textbf{RQ4}: How does the online cost and latency of SkillCDG compare with baselines?

\begin{table*}[t!]
\centering
\small
\caption{Main results across five test datasets. Columns report accuracy (ACC) and F1. Rows Qwen3-4B through DeepSeek-V4-Pro are backbone baselines; SkillCDG (4B) is the full method. The best and second-best results are highlighted through \textbf{boldface} and \underline{underline}, respectively, and the best results are also marked with a shaded cell.}
\label{tab:main}
\resizebox{\textwidth}{!}{%
\begin{tabular}{l|cc|cc|cc|cc|cc|cc}
\toprule
\multirow{2}{*}{Method}
  & \multicolumn{2}{c|}{Fulfillment}
  & \multicolumn{2}{c|}{AfterSales}
  & \multicolumn{2}{c|}{Merchant Support}
  & \multicolumn{2}{c|}{CompliBench}
  & \multicolumn{2}{c|}{LongBench}
  & \multicolumn{2}{c}{Overall} \\
& ACC & F1 & ACC & F1 & ACC & F1 & ACC & F1 & ACC & F1 & ACC & F1 \\
\midrule
\multicolumn{13}{c}{\textit{Full-SKILL}} \\
\midrule
Full-SKILL
  & 42.80 & 41.50 & 48.00 & 44.70 & 46.10 & 46.10
  & 21.60 & 23.10 & 6.28 & 4.94 & 32.96 & 32.07 \\
\midrule
\multicolumn{13}{c}{\textit{Retrieval-based}} \\
\midrule
BM25-RAG
  & 36.20 & 32.80 & 41.50 & 37.60 & 39.80 & 40.20
  & 18.60 & 20.40 & 19.30 & 21.10 & 31.08 & 30.42 \\
Dense-RAG
  & 39.80 & 36.10 & 44.20 & 40.30 & 42.60 & 43.10
  & 20.90 & 22.70 & 21.60 & 23.40 & 33.82 & 33.12 \\
GraphRAG
  & 42.30 & 38.50 & 47.10 & 43.00 & 45.20 & 45.90
  & 22.80 & 24.60 & 23.40 & 25.20 & 36.16 & 35.44 \\
LightRAG
  & 44.60 & 40.30 & 49.60 & 45.20 & 47.30 & 48.50
  & 24.10 & 25.80 & 24.70 & 26.40 & 38.06 & 37.24 \\
\midrule
\multicolumn{13}{c}{\textit{Backbone Baselines}} \\
\midrule
Qwen3-4B
  & 54.40 & 54.30 & 50.60 & 50.60 & 56.00 & 52.90
  & 23.60 & 24.80 & 26.10 & 27.60 & 42.14 & 42.04 \\
Qwen3-8B
  & 50.00 & 50.00 & 54.80 & \underline{54.80} & 50.80 & 50.00
  & 24.10 & 25.20 & 26.90 & 28.30 & 41.32 & 41.66 \\
DeepSeek-V4-Flash
  & 65.00 & 59.10 & 56.80 & \underline{54.80} & \underline{67.40} & 51.40
  & 24.50 & 24.30 & 26.40 & 25.80 & 48.02 & 43.08 \\
DeepSeek-V4-Pro
  & \underline{66.00} & \underline{71.10}
  & \cellcolor{green!15}\textbf{60.80} & \cellcolor{green!15}\textbf{54.81}
  & \underline{67.40} & \underline{58.43}
  & \underline{32.12} & \underline{35.30}
  & \underline{32.40} & \cellcolor{green!15}\textbf{35.08}
  & \underline{51.74} & \underline{50.94} \\
\midrule
\multicolumn{13}{c}{\textit{SkillCDG}} \\
\midrule
\textbf{SkillCDG (4B)}
  & \cellcolor{green!15}\textbf{71.08} & \cellcolor{green!15}\textbf{72.13}
  & \underline{58.73} & 52.93
  & \cellcolor{green!15}\textbf{68.32} & \cellcolor{green!15}\textbf{62.73}
  & \cellcolor{green!15}\textbf{33.85} & \cellcolor{green!15}\textbf{37.45}
  & \cellcolor{green!15}\textbf{39.83} & \underline{32.18}
  & \cellcolor{green!15}\textbf{54.36} & \cellcolor{green!15}\textbf{51.48} \\
\bottomrule
\end{tabular}}
\end{table*}

\subsection{Experimental Setup}

\subsubsection{Datasets}
We evaluate SkillCDG on three anonymized enterprise datasets and two public benchmarks.
Table~\ref{tab:dataset} summarizes the statistics for all five splits.

\textbf{Fulfillment} is a consumer-facing dataset from an online food-delivery platform,
covering customer service conversations governed by SOP documents for scenarios such as
delayed delivery and missing utensils.

\textbf{AfterSales} shares the same platform and SOP structure as Fulfillment but focuses on
high-value users, whose interactions are subject to additional escalation and compensation policies.

\textbf{Merchant Support} covers merchant-side operational consultations on the same platform,
governed by merchant-facing policy documents.

\textbf{CompliBench}~\citep{yang2026complibench} is a public compliance benchmark built from
general enterprise guidelines.

\textbf{LongBench}~\citep{bai2024longbench} is a bilingual multitask benchmark for long-context
understanding, spanning single-document QA, multi-document QA, summarization, few-shot learning,
and code completion.

\begin{table}[t]
\centering
\caption{Dataset statistics. \emph{Avg.\ SKILL len} is mean tokens per SKILL document.}
\label{tab:dataset}
\setlength{\tabcolsep}{4pt}
\resizebox{\columnwidth}{!}{
\begin{tabular}{lrrrrr}
\toprule
Dataset & Total & Compliant & Violating & Avg.\ SKILL len  \\
\midrule
Fulfillment      & 1{,}442 & 955  & 487  & 10{,}669\\
AfterSales       & 1{,}536 & 898  & 638  &  7{,}037  \\
Merchant Support &   870   & 630  & 240  &  5{,}031 \\
CompliBench & 2{,}710 & 827  & 1{,}883 & 739    \\
LongBench   &   2152   & 1360  & 792  & 24{,}674  \\
\bottomrule
\end{tabular}}
\end{table}

\subsubsection{Baselines}
We compare SkillCDG against two baseline groups.
\textbf{Full-SKILL}: the complete SKILL text is placed in context and a judge model produces a verdict directly without retrieval, defining the noise-level upper bound for context size.
\textbf{Retrieval-based}: BM25-RAG and Dense-RAG retrieve passages by lexical and semantic similarity; GraphRAG~\citep{edge2024graphrag} retrieves community-level summaries from a knowledge graph; LightRAG~\citep{guo2024lightrag} uses a lightweight entity-relation graph with relevance traversal.
Table~\ref{tab:main} also reports backbone baselines: \textbf{Qwen3-4B}, \textbf{Qwen3-8B}, \textbf{DeepSeek-V4-Flash}, and \textbf{DeepSeek-V4-Pro}.

\subsubsection{Model and Training Settings}
We use Qwen3-4B as the base model for training. The teacher model is Qwen3-30B-A3B. The retrieval stage allows at most four keywords per instance. Training and evaluation are performed on four A100 GPUs.

\subsubsection{Metrics}
\textbf{Accuracy} (ACC) is computed over all instances, with unparseable or retrieval-failed outputs counted as incorrect.
\textbf{Violation-F1} is reported on successfully parsed outputs with violation as the positive class.

\subsection{RQ1: Comparison with Baselines}

We first compare SkillCDG with the full-document, retrieval-based, and backbone baselines on the five test datasets. Table~\ref{tab:main} reports overall ACC and F1. The results show that SkillCDG (4B) achieves the best overall ACC (54.36) and F1 (51.48) among the evaluated methods.

In detail, Full-SKILL reaches 32.96 ACC and 32.07 F1, showing that the complete policy does not by itself yield accurate judgments when the context contains many inapplicable rules. Most retrieval-based methods improve over Full-SKILL, and LightRAG is the strongest at 38.06 ACC and 37.24 F1. It still trails SkillCDG by 16.30 ACC and 14.24 F1 points, since similarity search recovers surface-relevant rules but misses prerequisite rules that share no vocabulary with the scenario. Among the backbone baselines, Qwen3-4B and Qwen3-8B reach 42.14 and 41.32 ACC; the larger Qwen3-8B does not improve over Qwen3-4B, indicating that scale alone does not determine judgment quality. DeepSeek-V4-Flash reaches 48.02 ACC and 43.08 F1, and DeepSeek-V4-Pro is the strongest backbone baseline at 53.06 ACC and 51.28 F1. Although DeepSeek-V4-Pro is slightly better on AfterSales (60.80 vs 58.73 ACC), SkillCDG reaches its result with the 4B backbone and leads on the overall comparison and on the other four datasets.

To isolate the contribution of the retrieval stage, we compare SkillCDG with the frozen-retrieval variant in Table~\ref{tab:ablation}. The variant trails SkillCDG, so improving the judgment stage alone does not compensate for a frozen retrieval stage. These results indicate that the closure and joint training add to the backbone's judgment capability: the closure is the largest single contributor, and training the retrieval stage accounts for most of the gain over the frozen-retrieval variant. Appendix~\ref{app:implementation} provides the baseline and implementation details.

\subsection{RQ2: Ablation Study}

We ablate five variants of SkillCDG. The full model is compared with four variants: removing the dependency closure exposes only the keyword-seed rules to the judge; freezing retrieval at the SFT outputs trains only the judgment stage; applying OPD uniformly removes the entropy gate; and removing the teacher leaves the student with the graph-derived reward alone.
Table~\ref{tab:ablation} reports the results on Fulfillment and LongBench. The full model has the highest ACC on both splits. Its F1 is highest on Fulfillment, while on LongBench the uniform-OPD variant is slightly higher in F1 (33.04 versus 32.18). The four ablations rank in the same order by ACC: closure, retrieval-stage RL, teacher, entropy gate.

Removing the dependency closure causes the largest ACC drop, from 71.08 to 21.64 on Fulfillment and from 39.83 to 10.13 on LongBench.
Prerequisite rules that share no vocabulary with the scenario carry information that lexical matching does not recover.
Training the judgment stage alone does not compensate for freezing the retrieval stage.
The entropy gate leaves Fulfillment ACC nearly unchanged (71.08 to 69.41) and Fulfillment F1 nearly unchanged (72.13 to 70.32); on LongBench, uniform OPD is slightly higher in F1 (33.04 versus 32.18). Its effect is small on both metrics.
Together, the components produce a compact, conditionally complete rule set.

\begin{table}[t]
\centering
\caption{Ablation of SkillCDG components.}
\label{tab:ablation}
\setlength{\tabcolsep}{4pt}
\resizebox{\columnwidth}{!}{
\begin{tabular}{lcccc}
\toprule
& \multicolumn{2}{c}{Fulfillment} & \multicolumn{2}{c}{LongBench} \\
\cmidrule(lr){2-3}\cmidrule(lr){4-5}
Variant & ACC & F1  & ACC & F1  \\
\midrule
SkillCDG (full)                          & 71.08 & 72.13  & 39.83 & 32.18 \\
\quad w/o dependency closure (seeds only)& 21.64  & 21.53 & 10.13 & 20.63 \\
\quad w/o retrieval RL (frozen retrieval)& 53.19 & 43.03  & 25.50 & 23.73 \\
\quad w/o entropy gate (uniform OPD)     & 69.41 & 70.32 & 38.24 & 33.04 \\
\quad w/o teacher (joint RL only)        & 58.34 & 54.31 & 32.45 & 26.71 \\
\bottomrule
\end{tabular}}
\end{table}

\subsection{RQ3: Robustness to SKILL Length}

\begin{figure}[t]
\centering
\includegraphics[width=\linewidth]{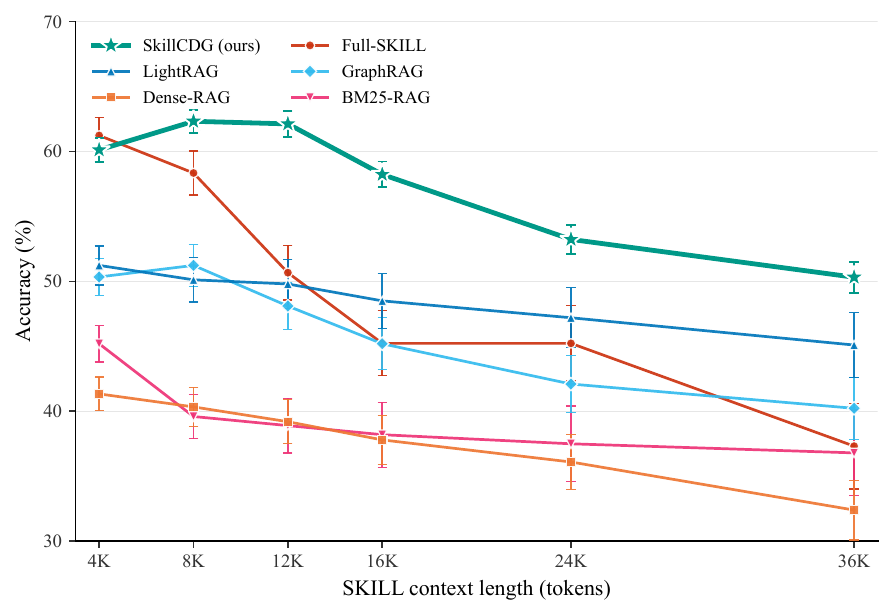}
\caption{Accuracy as the SKILL context length grows from 4K to 36K tokens on the length-controlled evaluation set.}
\label{fig:skill-scaling}
\end{figure}

To isolate the context-selection gap as a single variable, we hold the conversations and gold labels fixed and expand the SKILL documents loaded into the context from 4K to 36K tokens at six controlled lengths. The only variable is the number of inapplicable clauses added to the context, so any change in accuracy is attributable to length. We compare the same methods as RQ1: Full-SKILL, BM25-RAG, Dense-RAG, GraphRAG, LightRAG, and SkillCDG. Figure~\ref{fig:skill-scaling} reports accuracy at each length.

Full-SKILL carries every clause into the judge, so the added clauses accumulate as noise. Retrieval-based methods face the same pressure: the expanded document adds noise around the rules they retrieve. In SkillCDG, the added clauses are not reachable from the activated seeds, so they do not enter the closure and the judgment context stays close to the activated dependency chain. The experiment measures whether the gap between SkillCDG and the baselines grows as the context lengthens.

\subsection{RQ4: Cost and Latency Analysis}

Table~\ref{tab:cost} reports per-instance input tokens, output tokens, and judgment latency, measured on 4$\times$A100 GPUs with vLLM.
The retrieval stage adds a modest latency overhead relative to Full-SKILL (1.1--1.3$\times$), while the retrieval-based baselines have relative latency 1.5--1.9$\times$. It cuts prompt tokens by 76--86\% and improves ACC over Full-SKILL (Table~\ref{tab:main}).
Across the evaluated methods, SkillCDG gives the lowest prompt cost at higher ACC than Full-SKILL.

\begin{table}[t]
\centering
\caption{Online cost and latency per instance on Fulfillment and LongBench. Input and output tokens are mean prompt and completion tokens in thousands; Lat. is judgment latency, reported relative to Full-SKILL ($1.00\times$).}
\label{tab:cost}
\setlength{\tabcolsep}{4pt}
\resizebox{\columnwidth}{!}{%
\begin{tabular}{lcccccc}
\toprule
& \multicolumn{3}{c}{Fulfillment} & \multicolumn{3}{c}{LongBench} \\
\cmidrule(lr){2-4}\cmidrule(lr){5-7}
Method & Input (k) & Output (k) & Lat. & Input (k) & Output (k) & Lat. \\
\midrule
Full-SKILL      & 35.0 & / & \textbf{1.0} & 71.2 & \textbf{0.12} & \textbf{1.0} \\
BM25-RAG        & 21.8 & 0.38 & 1.8 & 37.3 & 0.34 & 1.7 \\
Dense-RAG       & 19.3 & \textbf{0.25} & 1.9 & 35.2 & 0.34 & 1.8 \\
GraphRAG        & 15.3 & \underline{0.28} & 1.5 & 29.3 & \underline{0.31} & 1.5 \\
LightRAG        & \underline{14.7} & 0.31 & 1.5 & \underline{20.5} & \underline{0.31} & 1.5 \\
SkillCDG (ours) & \textbf{8.5} & 0.31 & \underline{1.3} & \textbf{10.2} & \textbf{0.12} & \underline{1.1} \\
\bottomrule
\end{tabular}}
\end{table}

\section{Conclusion}
\label{sec:conclusion}

This paper presents SkillCDG, a framework for compliance detection under long SKILL documents.
The constraint dependency graph serves two purposes: it defines the judgment context as a dependency closure over mandatory prerequisites, and it supplies the non-emptiness gate that, together with verdict correctness, trains the retrieval stage without retrieval-level annotation.
Experiments on three enterprise datasets and two public benchmarks show that SkillCDG raises overall ACC from 32.96 for Full-SKILL to 54.36 and F1 from 32.07 to 51.48.

Two directions remain: closing the loop between closure failures and extractor training, and replacing the binary verdict reward with a signal that scores prerequisite-chain completeness.
Framing the applicable-rule set as an explicit, traceable artifact also provides a basis for audit trails beyond the verdict.

%%%%%%%%% AI Ethics / Responsible Research %%%%%%%%%
\section*{Ethical Considerations}
\label{sec:ethics}

\paragraph{Enterprise data.}
The enterprise datasets used in this work are proprietary business data provided under confidentiality agreements; all personally identifiable information has been removed prior to use.

\paragraph{Use of AI assistance.}
AI language models were used to assist with language polishing and writing during the preparation of this paper; all research design, experiments, and conclusions are the authors' own.

\paragraph{Misuse risk.}
This system is designed for transparent, auditable compliance detection; every verdict is traceable to a cited rule span, making the output suitable for legitimate oversight rather than adversarial exploitation.

\bibliographystyle{ACM-Reference-Format}
\bibliography{references}

\appendix

\section{CDG Construction Details}
\label{app:construction}

Algorithm~\ref{alg:construction} gives the procedure behind the three
construction steps in Section~\ref{sec:method-cdg-construction}. It processes a
SKILL document block by block, validates the extracted fields and spans, and
fuses rules across blocks. The output is the graph
$\mathcal{G}_j=(\mathcal{R}_j,\mathcal{E}_j)$ used by the retrieval stage.

\begin{algorithm}[t]
\caption{Block-wise CDG construction for one SKILL document}
\label{alg:construction}
\small
\begin{algorithmic}[1]
\STATE \textbf{Input:} SKILL document $D_j$ with name and description
\STATE \textbf{Output:} Graph $\mathcal{G}_j=(\mathcal{R}_j,\mathcal{E}_j)$
\STATE Split $D_j$ at Markdown headings into blocks $B_1,\dots,B_s$
\STATE Initialize $\mathcal{G}_j\leftarrow(\emptyset,\emptyset)$
\FOR{$B_\ell$ in document order}
  \STATE Extract atomic rules $\mathcal{R}_\ell$ from $B_\ell$
  \STATE Extract within-block \texttt{require} edges $\mathcal{E}_\ell$
  \STATE Validate fields, identifiers, endpoints, and source spans
  \STATE Rebase local rule identifiers to avoid collisions
  \IF{$\ell=1$}
    \STATE $\mathcal{G}_j\leftarrow(\mathcal{R}_\ell,\mathcal{E}_\ell)$
  \ELSE
    \STATE Predict the cross-block merge map $\mathcal{M}_\ell$
    \STATE Predict cross-block edges $\mathcal{E}_\ell^{\times}$
    \STATE Apply $\mathcal{M}_\ell$ to rules and edge endpoints
    \STATE Union the source spans of each merged rule
    \STATE Add the remaining rules in $\mathcal{R}_\ell$ to $\mathcal{R}_j$
    \STATE $\mathcal{E}_j\leftarrow
    \mathcal{E}_j\cup\mathcal{E}_\ell\cup\mathcal{E}_\ell^{\times}$
    \STATE Remove self-loops and duplicate edges
    \STATE $\mathcal{G}_j\leftarrow(\mathcal{R}_j,\mathcal{E}_j)$
  \ENDIF
\ENDFOR
\STATE Store $\mathcal{G}_j$ and index its name and description
\RETURN $\mathcal{G}_j$
\end{algorithmic}
\end{algorithm}

Cross-block fusion merges two rules only when their conditions and actions are
semantically equivalent. The merged rule keeps one identifier and the union of
both source spans, so traceability is preserved. Cross-block edges are added only
when the target is a mandatory prerequisite of the source, not when two rules are
only topically related. Self-loops and duplicate edges are removed after each
fusion step. Span validation follows the reject-rather-than-guess rule in
Section~\ref{sec:method-cdg-construction}.

Construction is offline. Its cost is dominated by the extractor calls, one per
block plus the fusion calls. The graph operations are linear in the number of
extracted rules and edges. The stored graph carries its name and description
index for document routing, and the rule graph itself for closure retrieval.

\section{Closure Computation}
\label{app:closure}

Algorithm~\ref{alg:closure} computes the closure in
Eq.~\eqref{eq:closure}. It processes the active documents separately and unions
the per-document closures, which avoids identifier collisions across SKILLs.
The queue is FIFO, and the visited set $\mathcal{C}$ doubles as the output and
prevents revisiting rules. Seeds and adjacency lists are processed in a fixed order so
that the same input produces the same closure. Here $\mathrm{tokens}(\mathcal{C})$
is the token length of the rendered context for $\mathcal{C}$.

\begin{algorithm}[t]
\caption{Dependency-closure retrieval}
\label{alg:closure}
\small
\begin{algorithmic}[1]
\STATE \textbf{Input:} Active documents $\mathcal{J}_x$ and seeds
$\mathcal{R}_q^{\mathrm{seed}}$
\STATE \textbf{Input:} Graphs $\mathcal{G}_j=(\mathcal{R}_j,\mathcal{E}_j)$ for
$j\in\mathcal{J}_x$
\STATE \textbf{Output:} Closure $\mathcal{C}$ and status
\STATE $\mathcal{C}\leftarrow\emptyset$
\FOR{each $j\in\mathcal{J}_x$ in a fixed order}
  \STATE $\mathcal{S}_j\leftarrow
  \mathcal{R}_q^{\mathrm{seed}}\cap\mathcal{R}_j$
  \STATE $Q_j\leftarrow\emptyset$, $\mathcal{C}_j\leftarrow\emptyset$
  \FOR{each $r\in\mathcal{S}_j$ in a fixed order}
    \STATE Add $r$ to $\mathcal{C}_j$ and enqueue $r$ into $Q_j$
  \ENDFOR
  \WHILE{$Q_j\neq\emptyset$}
    \STATE Dequeue $r_i$ from $Q_j$
    \FOR{each $(r_i,r_k)\in\mathcal{E}_j$ in a fixed order}
      \IF{$r_k\notin\mathcal{C}_j$}
        \STATE Add $r_k$ to $\mathcal{C}_j$ and enqueue $r_k$ into $Q_j$
      \ENDIF
    \ENDFOR
  \ENDWHILE
  \STATE $\mathcal{C}\leftarrow\mathcal{C}\cup\mathcal{C}_j$
\ENDFOR
\IF{$\mathcal{C}=\emptyset$}
  \RETURN $(\emptyset,\mathrm{EmptySeed})$
\ENDIF
\STATE Validate the source span $\mathcal{S}$ of every $r\in\mathcal{C}$
\IF{a span is missing}
  \RETURN $(\mathcal{C},\mathrm{SpanError})$
\ENDIF
\IF{a token budget $B$ is given and $\mathrm{tokens}(\mathcal{C})>B$}
  \RETURN $(\mathcal{C},\mathrm{BudgetExceeded})$
\ENDIF
\RETURN $(\mathcal{C},\mathrm{Ok})$
\end{algorithmic}
\end{algorithm}

\paragraph{Correctness.}
Let $\mathcal{R}_a=\bigcup_{j\in\mathcal{J}_x}\mathcal{R}_j$ and
$\mathcal{E}_a=\bigcup_{j\in\mathcal{J}_x}\mathcal{E}_j$. Every rule added to
$\mathcal{C}$ is either a seed or the target of a \texttt{require} edge whose
source is already in $\mathcal{C}$, so every rule in $\mathcal{C}$ is reachable
from a seed. Conversely, if a rule $r$ is reachable from a seed by a path of
\texttt{require} edges, the queue processes the path in order. When the
predecessor of $r$ on the path is dequeued, $r$ is added. Induction on the path
length gives $r\in\mathcal{C}$. Therefore
$\mathcal{C}=\mathrm{Cl}(\mathcal{R}_q^{\mathrm{seed}})$. Because the graphs
are document-scoped, every \texttt{require} path stays within one document, so
the union of the per-document closures equals the closure over the union graph.

\paragraph{Termination and complexity.}
Each rule enters a queue at most once, because a rule is enqueued only when it is
added to the visited set. The graph is finite, so the loop terminates even when
the graph contains cycles. Each rule in the closure is dequeued once, and each
outgoing edge of a dequeued rule is scanned once. The time complexity is
$O(|\mathcal{R}_a|+|\mathcal{E}_a|)$ in the worst case, and
$O(|\mathcal{C}|+|\mathcal{E}(\mathcal{C})|)$ when only the closure is
reachable, where $\mathcal{E}(\mathcal{C})$ is the set of edges leaving
$\mathcal{C}$. The space complexity is $O(|\mathcal{R}_a|)$ for the visited set
and the queue. Token counting adds $O(|\mathcal{C}|)$.

\paragraph{Edge cases.}
An empty seed set returns $\mathrm{EmptySeed}$ and the audit run is reported as a
retrieval error, as in Section~\ref{sec:method-closure}. The procedure treats an
empty seed set as an error rather than falling back to the full rule set.
A missing source span returns $\mathrm{SpanError}$, and no untraceable rule is
emitted. If the rendered closure exceeds a deployment's token budget $B$, the
procedure returns $\mathrm{BudgetExceeded}$ instead of truncating the closure,
since truncation can remove a prerequisite. The measured closure sizes in
Section~\ref{sec:experiments} are well below the corresponding full-document
contexts. Graphs are document-scoped. If a deployment has cross-document
prerequisites, the same procedure applies to a union graph with
document-prefixed rule identifiers.

\section{Case Study}
\label{app:case}

Algorithm~\ref{alg:closure} is instantiated on a Merchant Support violation
instance. We use a fixed judge from the evaluated backbones and the same judge
for both policy representations.

\subsection{Interaction}

\begin{quote}
\small\sloppy
\textbf{Conversation.} The merchant asks why an order was judged to violate the
rules. In the prior turn, the agent explains that the cancellation note says
``item inconsistent with the description / not prepared according to the order
note.'' The merchant objects that bundled-discount orders generally do not
support customer notes.

\textbf{Environment state.} The refund record and the cancellation record assign
responsibility to the merchant. The cancellation evidence states that the item
was inconsistent with its description or not prepared according to the order
note. The customer refund is 11.16 currency units, and the merchant's expected
income, which bounds the relevant deduction explanation, is 8.66.

\textbf{Candidate response.} ``You are right that bundled-discount orders do not
support customer notes. The basis for this judgment may be incorrect. I will
record it for further review, and a result is expected within 48 hours. For the
meal-loss claim, please contact the city manager. Would you like me to find the
contact information?''
\end{quote}

The response discounts the recorded cancellation evidence, omits the deduction
limit and the corrective recommendation, and introduces an unsupported 48-hour
review commitment.

\subsection{Extracted Graph and Closure}

The graph-construction model extracts six rules and three \texttt{require} edges
from the order-refund SKILL. Table~\ref{tab:case-rules} lists the rules.
Rules C032, C033, and C034 are directly applicable to the interaction. Their
closure adds C014, C050, and C007, so the six-rule closure is the context
presented to the judge.

\begin{table*}[t]
\centering
\small
\caption{Extracted rules and \texttt{require} edges for the Merchant Support
case. The right column gives the prerequisite of each rule.}
\label{tab:case-rules}
\begin{tabular}{p{0.07\textwidth}p{0.30\textwidth}p{0.36\textwidth}p{0.10\textwidth}}
\toprule
Rule & Condition & Action & Requires \\
\midrule
C032 & Merchant responsibility and refund explanation & State the refund
reason and cite responsibility evidence & C014 \\
C033 & Merchant responsibility and refund explanation & State the deduction
upper bound & C050 \\
C034 & Merchant responsibility and refund explanation & Provide an improvement
recommendation & C007 \\
C014 & Determining the refund reason or responsible party & Read both from
system signals & -- \\
C050 & Using the merchant-responsibility refund script & State merchant
responsibility and the deduction limit & -- \\
C007 & The refund is caused by the merchant & Explain honestly and provide
corrective guidance & -- \\
\bottomrule
\end{tabular}
\end{table*}

\subsection{Verdict}

Table~\ref{tab:case-verdict} compares the two policy representations. With the
full SKILL document, the judge returns a compliant verdict: the response
explains the refund and routes the merchant to a city manager, which reads as
helpful and within scope. With the closure, the judge returns a non-compliant
verdict: C033 requires the deduction upper bound and C034 requires a corrective
recommendation, and the response provides neither. The closure also brings in
C014, C050, and C007, which ground the responsibility evidence and the guidance
requirement. The case illustrates the difference between a context selected by
relevance alone and one that is closed under the prerequisites that the document
states.

\begin{table*}[t]
\centering
\small
\caption{Verdicts for the same instance under the two policy representations.
The closure verdict cites the rules whose requirements the response does not
meet.}
\label{tab:case-verdict}
\begin{tabular}{p{0.18\textwidth}p{0.14\textwidth}p{0.58\textwidth}}
\toprule
Policy context & Verdict & Reason \\
\midrule
Full-SKILL & Compliant & The response explains the refund and provides a
city-manager route without exceeding the SKILL scope. \\
Closure & Non-compliant & Rules C033 and C034 require the 8.66 deduction limit
and improvement advice. The response gives a review and city-manager routing
instead. \\
\bottomrule
\end{tabular}
\end{table*}

\section{Baseline and Implementation Details}
\label{app:implementation}

\paragraph{Full-SKILL.}
The Full-SKILL baseline concatenates the complete associated SKILL documents in
their original Markdown form. The judge performs relevance selection and
dependency reasoning internally. This baseline provides the complete source text
and defines the noise-level upper bound on context size.

\paragraph{BM25-RAG.}
BM25-RAG indexes the SKILL corpus at the passage level and retrieves passages by lexical overlap with the query. The retrieved passages are concatenated and passed to the same judge.

\paragraph{Dense-RAG.}
Dense-RAG encodes SKILL passages and the query with a dense retriever and retrieves the passages with the highest semantic similarity. The retrieved passages are concatenated and passed to the same judge.

\paragraph{GraphRAG.}
GraphRAG builds a knowledge graph from the SKILL corpus and aggregates community-level summaries. At inference it retrieves the summaries most relevant to the query and passes them to the same judge.

\paragraph{LightRAG.}
{\sloppy
LightRAG builds an entity-relation index over the same SKILL corpus. Documents
are segmented into 12{,}000-token chunks with 100-token overlap, and the index is
built once per dataset and shared across judges. Each entity and relation is
profiled with a textual description and linked source chunks. The online query
contains the associated SKILL names, the environment state, the conversation
history, and the candidate response, and requests complete condition-action
rules for the judge. Retrieval uses the keyword-oriented graph query mode with
\texttt{top\_k}=20, \texttt{chunk\_top\_k}=8, and a 6{,}000-token
retrieved-context cap.
\par}

\paragraph{Backbone baselines.}
Qwen3-4B, Qwen3-8B, DeepSeek-V4-Flash, and DeepSeek-V4-Pro are evaluated as backbone baselines on the same inputs, prompts, and response schema as SkillCDG. Only the backbone changes.

\paragraph{Shared controls.}
The judgment prompt, response schema, decoding settings, candidate responses,
and labels are fixed across methods. The parser validates the Boolean verdict
and the accompanying reason before aggregation. Online token counts include
keyword generation and judgment calls. Graph construction and index building are
offline.

\end{document}